\documentclass[12pt]{article}

\usepackage{arxiv}

\usepackage{amsmath}
\usepackage[utf8]{inputenc} 
\usepackage[T1]{fontenc}    
\usepackage{hyperref}       
\usepackage{url}            
\usepackage{booktabs}       
\usepackage{amsfonts}       
\usepackage{nicefrac}       
\usepackage{microtype}      
\usepackage{lipsum}

\usepackage{natbib,upgreek}
\usepackage{subcaption}
\usepackage{graphicx}
\newtheorem{remark}{Remark}
\newtheorem{definition}{Definition}

\usepackage{epstopdf}
\usepackage{footnote}

\usepackage{algorithmicx}
\usepackage{algorithm}
\usepackage[noend]{algpseudocode}
\usepackage{array}
\newcolumntype{C}[1]{>{\centering\arraybackslash}p{#1}}
\usepackage{multirow}
\usepackage{amsbsy}
\usepackage{tablefootnote}
\usepackage{gensymb}
\usepackage{xspace}

\usepackage{amssymb}

\def\bSig\mathbf{\Sigma}

\title{Harnessing the Power of Topological Data Analysis to Detect Change Points in Time Series}

\author{
  Umar ~Islambekov\\
  Department of Mathematical Sciences\\
  University of Texas at Dallas\\
  Richardson, TX 75080\\
  \texttt{umar.islambekov@utdallas.edu.edu} \\
   \And
  Monisha ~Yuvaraj \\
  Department of Mathematical Sciences\\
  University of Texas at Dallas\\
  Richardson, TX 75080\\
  \texttt{monisha.yuvaraj@utdallas.edu} \\
  \AND
  Yulia R.~Gel \\
  Department of Mathematical Sciences\\
  University of Texas at Dallas\\
  Richardson, TX 75080\\
  \texttt{ygl@utdallas.edu} \\
}

\begin{document}
\maketitle

\begin{abstract}

We introduce a novel geometry-oriented methodology, based on the emerging tools of topological data analysis, into the change point detection framework.
		The key rationale is that change points are likely to be associated with changes in geometry behind the data generating process. While the applications of topological data analysis to change point detection are potentially very broad, in this paper we primarily
		focus on integrating topological concepts with the existing nonparametric methods for change point detection. In particular, the proposed new geometry-oriented approach aims to enhance detection accuracy of distributional regime shift locations.
		Our simulation studies suggest that integration of topological data analysis with some existing algorithms for change point detection leads to consistently more accurate detection results.
		We illustrate our new methodology in application to the two closely related environmental time series datasets -- ice phenology of the Lake Baikal and the North Atlantic Oscillation indices, in a research query for a possible association between their estimated regime shift locations.
\end{abstract}


	\section{Introduction}\label{Intro}

	Change point detection (CPD) is broadly defined as the problem of identifying time instances within a time-ordered sequence of observations where changes in distribution of the data generating process (DGP) occur. If the distribution is known to belong to a parametric family, the changes are typically assumed to take place in the associated parameter value(s) such as mean or variance. If no such a distributional assumption can be made, nonparametric techniques are usually employed to find general changes in the distribution.
	
	The CPD problem naturally arises in the context of many applied problems such as image analysis, speech recognition, human activity analysis, medical condition monitoring, financial and environmental data analysis and modeling~\citep[see, e.g.,][]{Applications_image_analysis, Applications_speech_recog, CPD_human_activity, CPD_med_cond_monitor, Applications_financial, Applications_economic, CP_Alaska, Applications_environment}.
When analyzing environmental data, change points may arise due to several factors such as changes in the experimental design (e.g. change in equipment, procedure, location etc) or more crucially in the environment itself. Detecting change points is not only critical for investigating potential reasons that have caused them but also for proper modeling of the data and making correct inferences therefrom. Hence, assessing a given environmental time series data for presence of change points is an important step before the modeling stage~\citep{Review_3_CPD_climate_data}.
	
	A large class of CPD methods falls under a general framework that time series data are assumed to be piecewise homogeneous (e.g., stationary or independent and identically distributed) and the change points are defined to be the solutions of an appropriate optimization problem \citep{Review_1}. More specifically, this approach involves choosing a suitable cost function which measures homogeneity of a segment of data and optimizing a test statistic constructed using segment costs. A cost function is expected to be small on a data segment between two consecutive change points. In case the number of change points is unknown, a penalty term is incorporated as a measure of segment complexity. In this context, the CPD methods vary with respect to the choice of cost functions, the optimization algorithms and the assumption on the number of change points. The choice of cost functions is closely tied to the distributional assumptions behind the DGP and the types of changes that need to be detected. For example, in the parametric setting where the underlying distributions are of the same family, the likelihood function is commonly involved in formulating the associated optimization problem. On the other hand, nonparametric approaches are often based on techniques for density estimation. Furthermore, there are also nonparametric methods, utilizing rank statistics or pairwise Euclidean distances among observations and thereby bypassing the difficulty of estimating the density. More comprehensive surveys on CPD methods can be found, e.g., in \cite{Review_1, Review_2}.
	
	In this paper, we introduce a novel perspective into CPD, inspired by topological and geometric concepts. While most methods focus on detecting the distributional changes directly, we propose to analyze such changes indirectly -- through detecting changes in the corresponding topological/geometric structure underlying the DGP. In other words, we exploit the intrinsic relationship between distributional and topological/geometric changes associated with the process. To implement our idea, from a given times series data, we construct a new process whose observations encode the evolution of the local topology/geometry underlying the original times series. The measure of locality is controlled by the tuning parameter of window size. We then apply an existing nonparametric CPD method to the resulting time series of topological signatures. Our simulation studies show that the intermediate TDA-based step enhances the detection accuracy of changes occurring in the original times series (see Section \ref{Sec:Experiments} for more details).
	The choice of a nonparametric method can be explained by the fact that the distribution of the new observations is not known to belong to any parametric family. Under this approach, the cost function is effectively computed on the topological signatures instead of the original observations. To extract these signatures we resort to the mathematical tools provided by topological data analysis (TDA). One of the widely used TDA tools is the theory of \emph{persistent homology}. Having its roots in algebraic topology, persistent homology allows us to analyze data at multiple resolutions or scales in a unified way by keeping track of changes in the extracted topological signatures as the scale parameter increases \citep[]{Carlsson, TDAintro, CZ, Edelsbrunner}.
	
	The standard TDA framework for data analysis has a broad scope of applicability. Moreover, recent efforts aimed at developing fast algorithms to carry out TDA-related computations have facilitated the increase in its applications to diverse areas such as genomics \citep{B_cancer}, chemistry \citep{TDA_chem}, shape recognition \citep{S-recognition}, time series analysis \citep{TDA_ts}. To the best of our knowledge, the present paper is the first attempt to introduce the TDA-based techniques into the CPD problem.
	
	The paper is organized as follows. In Section \ref{Sec:TDA}, we present the applied aspects of TDA pertaining to our work. In Section \ref{Sec:Methodology}, we describe the methodology of our approach. Section \ref{Sec:Experiments} contains the results of extensive simulation studies performed using the proposed methodology. In Section \ref{Sec:Application}, we apply our method to a real-world data.
	
	\section{Topological data analysis (TDA)} \label{Sec:TDA}
	
	In this section, we provide a brief overview of TDA concepts from the applied perspective. The success of TDA in applications depends on addressing two interrelated challenges. The first one involves the input data which is assumed to lie in some metric space. In the context of persistent homology, the topological information is always extracted from the data via a medium. Typically this medium is a nested family, or a \textit{filtration}, of \emph{abstract simplicial complexes} (indexed by increasing scale values) built on the top of data points.
	\begin{definition}[Abstract simplicial complex]
		Let $X$ be a discrete set. An abstract simplicial complex is a collection $\mathcal{C}$ of finite subsets of $X$ such that if $\sigma\in\mathcal{C}$ then $\tau\in\mathcal{C}$ for all $\tau\subseteq\sigma$. If $|\sigma|=p+1$, then $\sigma$ is called a \emph{$p$-simplex}.
	\end{definition}
	
	A simplicial complex can be viewed as a higher dimensional generalization of graphs serving to approximate the geometric structure underlying data in a combinatorial way. In this context, a $0$-simplex can be identified with a point, a $1$-simplex with a line segment, $2$-simplex with a triangle, $3$-simplex with a tetrahedron and so on. Of the many kinds of simplicial complexes Vietoris-Rips is one of the most widely used in applications, most notably, due to its desired computational properties.
	
	\begin{definition}[Vietoris-Rips complex]
		Let $X$ be a point cloud lying in a metric space. A Vietoris-Rips complex on $X$ at scale $\epsilon\geq 0$,
		denoted by $VR_\epsilon$, is an abstract simplicial complex
		whose $p$-simplices, $p=0,\ldots,d$, consist of points which are pairwise within $\epsilon$ distance of each other.
		Here, $d$ is called the dimension of the complex. 		
	\end{definition}
	
	The second challenge in TDA has to do with the output of persistent homology: how can it be used to better understand and make inferences from the data? The two common outputs of persistent homology are a \emph{persistence diagram} (PD) and a set of sequences of \emph{Betti numbers}. A PD is a set of paired scale values corresponding to the appearance and disappearance of topological features (connected components, loops, voids, etc.) as the scale parameter increases.
	
	One major drawback of dealing with a PD is that often it cannot be easily incorporated into statistical or machine learning methods for further analysis of the data. On the other hand, the sequences of Betti numbers which encode the counts of the topological features as a function of the scale parameter, suffer from no such limitation and can easily be used in a variety of settings such as statistical modeling, machine learning and functional data analysis.
	
	\begin{definition}[Betti numbers]
		Betti-$p$ number $\beta_p$, $p\in \mathbb{Z}^{+}$, of a simplicial complex is the number of  $p$-dimensional holes in the complex. For example, $\beta_0$ is the number of connected components; $\beta_1$ is the number of loops; $\beta_2$ is the number of voids, etc. Betti-$p$ number of the VR complex at threshold $\epsilon$ is denoted by $\beta_p(\epsilon)$.
	\end{definition}
	
	In our methodology, we primarily focus on Betti sequences as a form of extracted topological signatures. Working with Betti sequences ensures that the observations of the TDA-derived times series, on which the CPD is performed, belong to the Euclidean space as the original observations do. If the associated cost function involves only pairwise distances between observations, we may take PDs as an alternative to Betti sequences since the space of PDs can be endowed with a notion of distance, such as bottleneck or Wasserstein \citep{Comp_top}.
	
	\section{Methodology}\label{Sec:Methodology}
	In this paper we consider a CPD framework under which the number of change points is assumed to be known. Suppose a time-ordered sequence of observations $y_1,y_2,\ldots,y_T$ are realizations of random variables $Y_1,Y_2,\ldots,Y_T$ which undergo distributional changes at unknown change points $\tau_1,\tau_2,\ldots,\tau_k$: $$ Y_i \sim \left\{
	\begin{array}{ll}
	F_1 & 1\leq i<\tau_1, \\
	F_2 & \tau_1\leq i<\tau_2, \\
	\cdots\\
	F_{k+1} & \tau_k \leq i \leq T
	\end{array}
	\right.
	$$
	Since TDA-extracted topological signatures such as Betti sequences and PDs are translation invariant \citep[see e.g.,][]{TDAintro}, we focus on detecting changes in variance and distributional changes assuming no shift in mean. Our goal is to estimate the unknown change points $\tau_1,\tau_2,\ldots,\tau_k$. Observations between two consecutive change points are commonly assumed to be independent. The case of autocorrelated observations can be handled by first pre-whitening the underlying process and then performing CPD on the model residuals.
	
	The main idea of our methodology is to identify the distributional changes in DGP through locating the associated changes in the topological/geometric structure of the data. The approach proposed in this paper works in conjunction with an existing CPD method, belonging to a general class of nonparametric techniques (see Section \ref{Sec:Conclusion} for future directions with regard to using TDA-based techniques in CPD). More specifically, we apply the selected CPD method to a new time series wherein the observations reflect 
	the local topological/geometric structure corresponding to the original time series. In turn, to construct such a time series of topological signatures, we employ the tools of TDA. Our method can be schematically represented as follows:
	$$
	\{y_t\}_{t=1}^{T}\subseteq \mathbb{R}^d
\rightarrow
{\hbox{TDA}}\{\tilde{y}_t\}_{t=1}^{T-w+1}\subseteq \mathbb{R}^m 
\rightarrow
{\hbox{CPD method}}
	\tilde{\tau_1},\tilde{\tau_2},\ldots,\tilde{\tau_k},
	$$
	where $w$ is a tuning parameter of window size (see details below) and $\tilde{\tau_1},\tilde{\tau_2},\ldots,\tilde{\tau_k}$ are the change point estimates. Below is a step-by-step description of the algorithm followed by its pseudo-code.
	
	\begin{enumerate}
		\item Given time series data $y=\{y_t\}_{t=1}^{T}\subseteq \mathbb{R}^\ell$. Slide a window of size $w$ along the (normalized) list $y$ and form a sequence of point clouds $\mathbb{X}=\{X_t\}_{t=1}^{T-w+1}$ where $X_t=\{y_t,y_{t+1},\ldots,y_{t+w-1}\}$.
		\item Fix an increasing sequence of scales: $0<\epsilon_1<\epsilon_2<\ldots<\epsilon_n$. Construct the corresponding filtration Vietoris-Rips complexes built on point clouds $X_t$:  $VR_{\epsilon_1}^{(t)}\subseteq VR_{\epsilon_2}^{(t)}\subseteq \ldots \subseteq VR_{\epsilon_n}^{(t)}$, and compute topological signatures in the form of the sequences of Betti numbers: $\boldsymbol\beta_{p}^{(t)}=(\beta_{p}^{(t)}(\epsilon_1),\beta_{p}^{(t)}(\epsilon_2),\ldots,\beta_{p}^{(t)}(\epsilon_n))$, $p=0,1,\ldots,d$. By default, $n=50$ and $\epsilon_i=0.01(i-1)$, $i=1,\ldots,n$.
		\item Using a suitable dimension reduction technique, map $\{ \boldsymbol\beta_{p}^{(t)}\}_{t=1,p=0}^{T-w+1,d}$ into a reduced $m$ dimensional space ($m<<n$) to construct a new time series $\tilde{y}=\{\tilde{y}_t\}_{t=1}^{T-w+1}\subseteq \mathbb{R}^m$. In this paper, we set $d=0$ and use the standard principal component analysis (PCA) to reduce the dimension (see also Remark \ref{why_betti-0_and_pca}). \label{PCA_step}
		\item Apply the selected CPD method to the TDA-derived time series $\tilde{y}=\{\tilde{y}_t\}_{t=1}^{T-w+1}$ to estimate the change point locations.
	\end{enumerate}
	
	\begin{algorithm}
		\caption{TDA change point algorithm}
		\label{euclid}
		\begin{algorithmic}[]
			\Procedure{TDA-Based CPD}{Time series $y=\{y_t\}_{t=1}^{T}$, Window size $w$, $\#$ PCA coordinates $m$, Increasing scale sequence $\epsilon[ ]$, Nonparametric CPD method $cpm$}
			\State $y$ = Normalize($y$)
			\For{$t$ = 1 to T}
			\State{$X[t]$ = $y$[$t$ to $t$+$w$]}
			\For{$i$ = 1 to $length(\epsilon[])$}
			\State {$VR[t,i]$ = VR complex ($X[t]$,$\epsilon[i]$)}
			\State {$\beta_0[t,i]$ = Betti-0 number($VR[t,i]$)}
			\EndFor
			\EndFor
			\State {$\tilde{y}$ = Perform PCA($\beta_0[]$)[1 to $m$]}
			
			
			\State {$ChangePoint$ = $cpm$($\tilde{y}$)}
			\State \textbf{return $ChangePoint$}
			\EndProcedure
		\end{algorithmic}
	\end{algorithm}
	
	\begin{remark}
		As a data preprocessing step, normalization is performed so that each $y_t$ lies in $[-{1}/{2},{1}/{2}]^\ell$. This step allows us to choose a sequence of increasing scales $\{\epsilon_i\}_{i=1}^n\subseteq[0,\sqrt{\ell}]$ in a way independent of the original scale the observations are measured in. For all the studies in this paper, we use the default values of $\{\epsilon_i\}_{i=1}^n$.
	\end{remark}
	
	\begin{remark}
		If the selected CPD method applies only to univariate data, then we choose $m=1$ in step \ref{PCA_step} so that the new TDA-derived time series is also univariate. In this case the original time series may even be multivariate.
	\end{remark}
	
	\begin{remark}\label{why_betti-0_and_pca}
		Since in applications the window size $w$ is taken to be relatively small, we do not normally observe topological features of dimension greater than zero (such as loops and voids) associated with $X_t$. Hence, we confine ourselves with Betti-0 sequences $\{\boldsymbol\beta_{0}^{(t)}\}$. Moreover, since the elements of 
	$\boldsymbol\beta_{0}^{(t)}$ take at most $w$ distinct values, PCA is performed to decrease information redundancy and to reduce computational complexity.
	\end{remark}
	
	
	\section{Simulation studies}\label{Sec:Experiments}
	\subsection{Experimental Setup}
	The proposed methodology is tested against the following competing approaches:
	\begin{itemize}
		\item \textbf{E-Divisive} \citep{E-divisive}: The E-Divisive method estimates multiple change points in multivariate time series data by recursively applying the energy statistic-based procedure \citep{Energy_stat} to find a single change point.
		\item \textbf{CvM CPM} \citep{CM}: The method integrates the omnibus (distribution-free) Cramer-von Mises test statistic into the change point analysis framework for a univariate time series.
	\end{itemize}
	
	The proposed geometry-oriented approach is not restricted to integration with E-Divisive and CvM CPM, and these CPD methods are selected for illustrative purposes. Both methods are nonparametric in nature and designed to detect general distributional changes. In principle, our methodology can be applied to any nonparametric CPD method. When the two selected CPD methods are applied in combination with our TDA-based methodology, we refer to them as E-Divisive+TDA and CvM CPM+TDA respectively.
	
	Our simulation set-up aims to cover two scenarios: change in variance and change in the shape of the distribution assuming no mean shift. Since topological signatures are shift invariant, our method is not sensitive to changes is mean while the variance remains constant. 
	In such a situation the geometric/topological structure underlying a point cloud does not change and hence in our simulation studies we do not consider changes in mean.
	
	
	As evaluation metrics, we use Mean Absolute Error (MAE). The number of Monte Carlo simulations is set to 1000 per a considered scenario. While the proposed CPD methodology is applicable to detection of multiple change points, in our current framework, we focus on identifying a single change point.
	
\begin{figure}[h]
		\centering
		\includegraphics[scale=0.35]{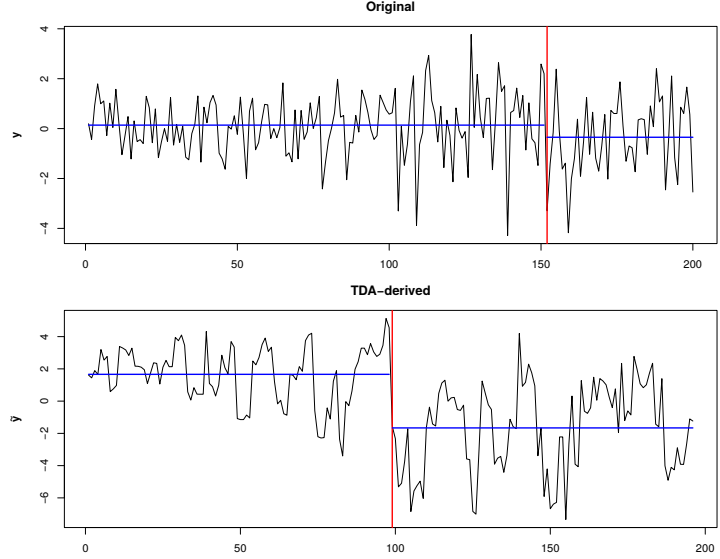}
		\caption{Top: simulated time series of independent normal observations with increase in variance at the midpoint. Bottom: the corresponding TDA-derived time series with $w=5$ and $m=1$. The red vertical lines correspond to the estimated change points using the E-divisive method. The blue horizontal lines are the means of the observations before and after the estimated change points.}
		\label{ts_plots}
	\end{figure}

\begin{figure}[h]
		\centering
		\includegraphics[scale=0.8]{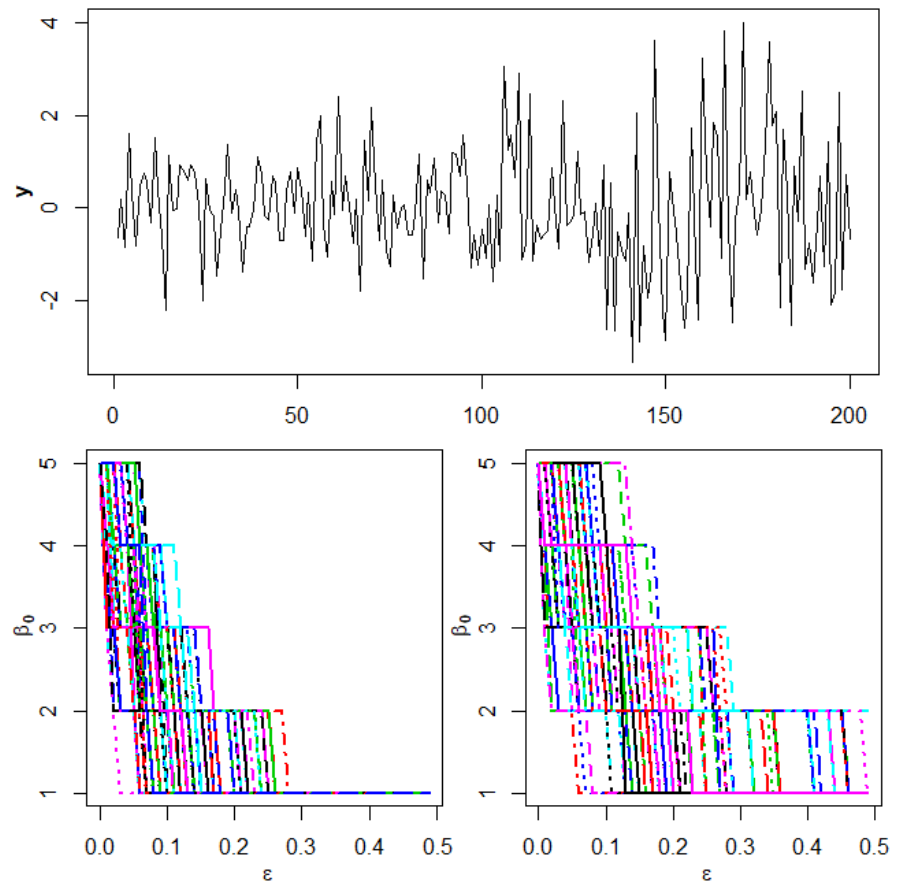}
		\caption{Top: simulated time series of independent normal observations with increase in variance at the midpoint. Bottom left: plots of Betti-0 sequences $\{\boldsymbol\beta_{0}^{(t)}\}_{t=1}^{100}$ corresponding to $\{X_t\}_{t=1}^{100}$. Bottom right: plots of Betti-0 sequences $\{\boldsymbol\beta_{0}^{(t)}\}_{t=101}^{196}$ corresponding to $\{X_t\}_{t=101}^{196}$. Window size $w=5.$}
		\label{Betti_curves}
	\end{figure}

	Figure \ref{ts_plots} illustrates the TDA-derived times series when the number of PCA coordinates $m$ is set to 1. In this example, the original observations are sampled from a normal distribution whose variance increases from 1 to 3 at the midpoint. From Figure \ref{ts_plots} we observe that the variance increase in $y$ not only translates into a similar increase in variance but also a marked mean shift in $\tilde{y}$ by the TDA procedure (see also Figure \ref{Betti_curves} for the plots of Betti-0 sequences $\{\boldsymbol\beta_{0}^{(t)}\}$ before and after the change point).

	We perform experiments with time series of length 200, 500 and 1000 data points. For the sake of room and to highlight performance of the proposed approach on moderate sample sizes, we report only the results for series of length 200. Studies on time series of length 500 and 1000 show analogous results and are available from the authors.
	
	Furthermore, additional parameters for the new TDA-based method, namely, the window size $w$ and the number of PCA coordinates $m$, are selected as follows. Based on the extensive simulations, we find that window size values in the range of 5\%-10\% of the time series length yield the best performance and can be set as a rule of thumb. The selection of the number of PCA coordinates is dependent on the type of CPD methodology used. CPD methods that are only applicable to univariate time series, in our case CvM CPD, the number of PCA coordinates is set to 1, while methods that work with multivariate data such as E-divisive offer more flexibility in this regard. However, our simulation studies show that the impact of the number of PCA components on the performance is very marginal.
	
	The E-Divisive procedure requires the minimum number of observations between change points which is set to 30 (default).
	
	All computations are performed using the statistical software R by \cite{R}; E-divisive is available within the {\fontfamily{cmtt}\selectfont ecp} package  \citep{james2013ecp} and {\fontfamily{cmtt}\selectfont cpm} package  \citep{ross2015parametric} is leveraged for the CvM CPM algorithm. The TDA methodology is implemented utilizing {\fontfamily{cmtt}\selectfont TDA} package \citep{fasy2014introduction}.
	
\subsubsection{Change in variance}
	\begin{table}[hbt!]
	\caption{Performance of TDA based methodology on detecting changes in variance}
\centering
 \begin{tabular}{@{}l c c c c c c@{}}\toprule
 & &  \multicolumn{4}{c}{Mean Absolute Error (MAE)} &\\
 \cmidrule(lr){3-6}
  & &  \multicolumn{2}{c}{CvM CPM} & \multicolumn{3}{c}{E-Divisive}  \\
 \cmidrule(lr){3-4}\cmidrule(lr){5-7}
   \multicolumn{1}{l}{Distributions} & \multicolumn{1}{{C{1.5cm}}}{Window Size} &  \multicolumn{1}{{C{1.5cm}}}{TDA-based} & \multicolumn{1}{{C{1.5cm}}}{Original} & \multicolumn{1}{{C{1.5cm}}}{TDA-based} & \multicolumn{1}{{C{1.5cm}}}{Original} &
   \multicolumn{1}{{C{2.5cm}}}{\# PCA Coordinates} \\
   \midrule
  \multirow{4}{6em}{$N(0,1) \rightarrow N(0,2)$}
  & 5  & 20.56 & 96.40 & 17.94 & 30.99& 3 \\
 &5   & 20.56 & 96.40 & 17.93 & 30.99 & 5\\
 & 10  &  17.68 & 96.40 &16.52 & 30.99 & 3 \\
 & 10   & 17.68 & 96.40 & 16.56 & 30.99& 5\\
 \midrule
  \multirow{4}{6em}{
  $\mathcal{N}(\pmb{\mu,\Sigma_\mathrm{1}})
 \rightarrow
 \mathcal{N}(\pmb{\mu,\Sigma_\mathrm{2}})$
}
& 5 & \multicolumn{2}{c}{\multirow{4}{4em}{\small{CvM CPM does not support multivariate time series}}} & 7.34 & 13.52 & 3  \\
 & 5  &   & & 7.28 & 13.52 &5 \\
 & 10  &   & & 5.01 & 13.52 &3 \\
 & 10  &   & & 5.06 & 13.52 &3 \\
 \midrule
 \multirow{4}{5em}{$Pois(1)-1 \rightarrow Pois(2)-2$}
 & 5  &  23.07 & 23.91 & 16.25 & 28.29 &3\\
 & 5  &  23.07 & 23.91 & 16.40 & 28.29 &5\\
 & 10  &  17.33 & 23.91 & 14.17 & 28.29 &3\\
 & 10  &  17.33 & 23.91 &14.17 & 28.29 & 5\\
 \midrule
 \multirow{4}{4em}{ARMA(1,1) Error $\epsilon_t$: $N(0,1) \rightarrow N(0,2)$ }
 & 5  &  22.24 &72.12 & 17.94 & 35.59 &3 \\
 & 5  &  22.24 & 72.12 &17.80 & 35.59 & 5\\
 & 10  &  20.91 & 72.12 & 18.89 & 35.59 &3\\
 & 10  &  20.91 & 72.12 & 19.03 & 35.59 &5\\
 \midrule
  \multicolumn{7}{l}{\footnotesize{$\pmb{\mu} =
  \begin{bmatrix}
0 \\
0 \\
0
\end{bmatrix}
\pmb{\Sigma_\mathrm{1}} = \begin{bmatrix}
1 & 0 & 0\\
0 & 1 & 0\\
0 & 0 & 1
\end{bmatrix} , \pmb{\Sigma_\mathrm{2}} = \begin{bmatrix}
1 & 0.9 & 0.9\\
0.9 & 1 & 0.9\\
0.9 & 0.9 & 1
\end{bmatrix}$}, ARMA(1,1): $x_t=0.4\cdot x_{t-1}+\epsilon_t+0.5\cdot \epsilon_{t-1}$} \\
\bottomrule
 	\label{Table 1}
 \end{tabular}
\end{table}

	We start from comparing performance with changes in variance for Normal, Multivariate Normal, Poisson distributions and for the error term distribution of ARMA(1,1) model. The Poisson distribution is mean-adjusted to compare between changes in variance and not changes in mean.
	
	The results in Table~\ref{Table 1} indicate that when E-divisive and CvM CPM methods are combined with the TDA-based methodology, there is a decrease in MAE ranging from a 42\% to 91\% compared to the baseline performance of the two methods.
	Moreover, we observe that overall E-divisive+TDA performs substantially better than CvM CPM+TDA. We suspect that this phenomenon can be explained by a higher flexibility of the E-divisive method with respect to CPM. For CvM CPM, we use associated empirical distributions, while for E-Divisive, the test statistic is defined in terms of pairwise distances among observations. Remarkably, the TDA methodology provides the best results in the multivariate case. This observation might be potentially explained by
	a richer geometrical structure of the data in higher dimensions, and hence, a higher utility of TDA tools for detecting structural changes.
	
	When a time series is generated by ARMA(1,1), the observations are no longer independent -- a key assumption for both E-Divisive and CvM CPM. Due to this reason, the two methods perform poorly on the original observations. In turn, E-divisive+TDA and CvM CPM+TDA appear to be less sensitive to serial correlations and deliver more stable and accurate performance. With regards to the robustness of results, in 88\% of the experiments the variance of the MAEs of E-divisive+TDA and CvM CPM+TDA is lower than that of the baseline methods.
	
	When the observations are sampled from a normal distribution whose variance increases by one unit, it is more appropriate to use CvM CPM with the standard Barlett's test for homogeneity of variances rather than the nonparametric Cramer-von  Mises test. With Barlett's test, the MAE drops from 96.4 to 36.7 which interestingly is still higher than the MAEs from the TDA procedure (see the first row in Table \ref{Table 1}).
	
	\subsubsection{Distributional Changes}
	
	\begin{table}[hbt!]
		\caption{Performance of TDA-based methodology on detecting distributional changes}
		\centering
		\begin{tabular}{@{}l c c c c c c@{}}\toprule
			& &  \multicolumn{4}{c}{Mean Absolute Error (MAE)}&\\
			\cmidrule(lr){3-6}
			& & \multicolumn{2}{c}{CvM CPM} & \multicolumn{3}{c}{E-Divisive}   \\
			\cmidrule(lr){3-4}\cmidrule(lr){5-7}
			\multicolumn{1}{l}{Distributions} & \multicolumn{1}{{C{1.5cm}}}{Window Size (\%)} &  \multicolumn{1}{{C{1.5cm}}}{TDA-based} & \multicolumn{1}{{C{1.5cm}}}{Original} & \multicolumn{1}{{C{1.5cm}}}{TDA-based} & \multicolumn{1}{{C{1.5cm}}}{Original} &
			\multicolumn{1}{{C{2.5cm}}}{\# PCA Coordinates} \\ \midrule
			\multirow{4}{6em}{$N(0,1) \rightarrow t(df=4)$}
			& 2.5  &  42.24 & 99.15 & 26.68 & 36.72 &3  \\
			& 2.5 &  42.24 & 99.15 & 26.35 & 36.72 & 5 \\
			& 5  &  30.52 & 99.15 &23.30 & 36.72 &3 \\
			& 5  &  30.52 & 99.15 & 23.55 & 36.72 &5\\
			\midrule
			\multirow{4}{6em}{
				$\mathcal{N}(\pmb{\mu,\Sigma_\mathrm{1}})
				\rightarrow
				t_\mathrm{2}(\pmb{\mu,\Sigma_\mathrm{2}})$
			}
			& 2.5  &   \multicolumn{2}{c}{\multirow{4}{4em}{\small{CvM CPM does not support multivariate time series}}} & 12.29 & 24.28 &3   \\
			& 2.5  &   &  & 12.59 & 24.28 &5\\
			& 5  &   & & 11.16 & 24.28 &3 \\
			& 5 &   & & 11.17 & 24.28 &5  \\
			\midrule
			\multirow{4}{4em}{$N(0,1) \rightarrow Laplace(\sqrt{0.5})$}
			& 2.5  &  56.70 &98.63 &32.88 & 37.20 & 3\\
			& 2.5  &  56.70 & 98.63& 32.95 & 37.20 &5 \\
			& 5  &  46.43 & 98.63& 31.44 & 37.20 &3 \\
			& 5  &  46.43 & 98.63&31.22 & 37.20 & 5 \\
			\midrule
			\multicolumn{7}{l}{\footnotesize{$\pmb{\mu} = \begin{bmatrix}
					0 \\
					0 \\
					0
					\end{bmatrix}
					\pmb{\Sigma_\mathrm{1}} = \begin{bmatrix}
					1 & 0 & 0\\
					0 & 1 & 0\\
					0 & 0 & 1
					\end{bmatrix} , \pmb{\Sigma_\mathrm{2}} = \begin{bmatrix}
					1 & 0.9 & 0.9\\
					0.9 & 1 & 0.9\\
					0.9 & 0.9 & 1
					\end{bmatrix}$}}
			\\ \bottomrule
			\label{Table 2}
		\end{tabular}
	\end{table}
	
	Distributional changes are studied between the following pairs of distributions: Normal and Student's $t$, Multivariate Normal and Multivariate Student's $t$, Normal and Laplace. In all three settings, the means are held fixed.
	
	The results in Table~\ref{Table 2} indicate that the new geometry-oriented CPD approaches substantially outperforms the baseline methods in all considered cases of distributional changes. The minimal gain via integration of TDA to CPD (i.e., 11\% decrease in MAE for E-divisive+TDA over E-divisive) is observed when the distribution shifts occur for the pair of $N(0,1)$ to $Laplace(\sqrt{0.5})$ distributions. Since both distributions have the same mean and variance, such a result is not surprising. Similarly to the case of change in variance, the best performance of the TDA-enhanced approach is again achieved for the multivariate case. Finally, under all simulation settings the number of PCA coordinates yields little to no effect on the performance of E-divisive+TDA and CvM CPM+TDA.
	
	\section{A case study on change point detection for the Lake Baikal and the North Atlantic Oscillation index datasets}\label{Sec:Application}
	
	Lake Baikal is estimated to be over 25 million years old and is the deepest freshwater lake on earth (1,650 m maximum depth) and the largest by volume, containing 20\% of all liquid
	fresh water on the surface of the earth. 
	Baikal spans 4\degree latitude which exposes it to a wide range of climactic conditions. Although thawing processes are extremely complex and involve many variables, air temperature appears to be by far the most important of these variables, and air temperature alone is often able statistically to explain 60\% - 70\% of the variance in the ice phenology of the Lake Baikal. Hence, not only historical air temperature data can be used to estimate the time of break-up of a lake, but also historical observations of the timing of break-up can be employed as proxy data for integrated local and regional air temperatures or temperature changes. Furthermore, as some studies indicate, remote sensing of ice cover can be used for estimating air temperatures in sparsely populated areas~\citep{livingstone1999ice}.
	Such interrelationships of local and regional climate and the Lake Baikal ice phenology motivate us to investigate the dynamics and regime shifts of the two environmental processes -- ice break-up dates for Lake Baikal and the North Atlantic Oscillation (NAO) index.
	
	\subsection{Datasets}	
	The dates of ice break-up on southern Lake Baikal have been
	registered continuously at the same point of observation, the
	Listvyanka limnological station (51\degree 85'29N, 104\degree85'19E) since 1869 \citep{shimaraev1993deep}. The term break-up date as used here refers to the first day on which the lake opposite the observation point is observed to be ice-free. The North Atlantic Oscillation (NAO) is one of the major atmospheric modes of variability in the Northern Hemisphere exerting strong influence on its climate, especially in winter \citep{livingstone1999ice}. A traditional way to define the NAO is as the normalized difference in pressure between a station on the Azores and another on Iceland \citep{NAO_data}.
	
	The datasets consist of yearly observations spanning the period from 1869 to 1996 (the yearly NAO index is the average of monthly measurements) \footnote{Ice phenology data for the Lake Baikal are kindly provided by Claude Duguay, University of Waterloo, Canada} \footnote{The NAO index dataset is available at https://crudata.uea.ac.uk/cru/data/nao}.
	
	\subsection{Methods and Results}
	
	Based on the simulation study results, we choose E-Divisive as a primary CPD method for the analysis of the Baikal ice break-up and the NAO index datasets. For comparative purposes, we also report the CPD results when E-Divisive is applied alone without TDA.
	\begin{figure}[h]
		\centering
		\includegraphics[scale=1]{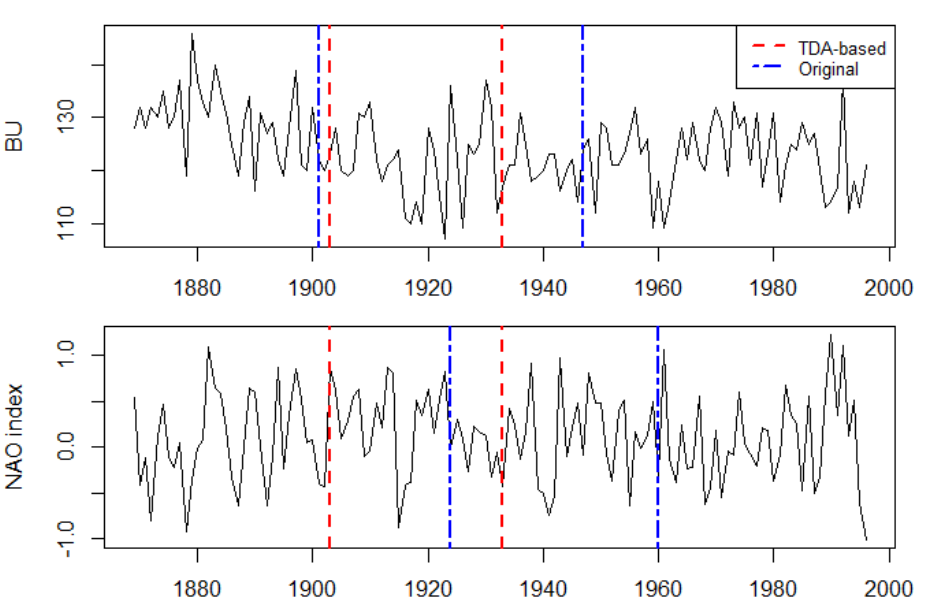}
		\caption{Change points for Ice break up and NAO index}
		\label{fig:both}
	\end{figure}
Note that the observations of the two time series are serially correlated and their dependence structure can be approximated by autoregressive models AR(6) and AR(1), respectively \citep{BaikalTrendshift}. The CPD methods are then applied to the corresponding model residuals. Earlier studies of the Baikal ice break-up data for change point analysis suggest the presence of two change points \citep[see e.g.,][]{BaikalTrendshift}. Hence, we focus on identifying the locations of these two change points. To assess the robustness of the TDA-based approach against the violations of method assumptions, such as independence of observations, we also perform CPD on the original unprocessed data and also after the first order differencing. When TDA is integrated into the CPD process, the results suggest that under all three different data pre-processing scenarios, the estimated change points for the Baikal ice break-up data essentially agree and fall within intervals 1899-1903 and 1932-1933 (see Figure~\ref{fig:both}, Tables~\ref{Table 3} and \ref{Table 4}). Similar intervals are observed for the NAO index data, indicating a strong association between the two time series data in term of their change point locations.
	
	Another compelling comparison can be made with the beginning of ice formation in the Baikal area which typically occurs before January 15. Interestingly, in winters of 1898-–1899 and 1931–-1932, ice formation in Listvyanka started as late as early February \citep{Lake_Baikal} and these two periods almost exactly coincide with the found change point intervals for the Baikal ice break-up dates.
	
	On the other hand, our results somewhat differ from those in \cite{BaikalTrendshift} where the same time series is analyzed for CPD using sieve-bootstrap technique to remedy the problem of serial correlations among the observations. Years 1886 and 1924 are found to be candidate years for change points. There is approximately a ten year difference between the two change point results. The divergence of results seems to indicate the changes did not happen abruptly but rather extended over the period of ten years.
	
	Based on these findings, we are likely to conclude that regime shifts in ice phenology of the Lake Baikal and NAO are indeed aligned.
	In turn, analysis of structural breaks in climate networks formed
	by ice phenology of the Lake Baikal, NAO
	and other climate indices
	may yield a deeper insight into teleconnection patterns and interrelationships of regional and global climate.
	
	\begin{table}[hbt!]\centering
		\caption{Estimated change points for ice break-up dates for Lake Baikal and the NAO index}
		\begin{tabular}{@{}c cc @{}}\toprule
			& Residuals of AR(6)& Residuals of AR(1)\\
			\cmidrule(lr){2-3}
			& BU  Baikal & NAO \\
			\midrule
			E-divisive & 1901 1947 & 1924 1960 \\
			E-divisive+TDA &  1903 1933 & 1903 1933 \\
			\bottomrule
		\end{tabular}\label{Table 3}
	\end{table}
	
	\begin{table}[hbt!]\centering
		\caption{Estimated change points for ice break-up dates for Lake Baikal and the NAO index}
		\begin{tabular}{@{}c cc cc@{}}\toprule
			& \multicolumn{2}{c}{No pre-processing} & \multicolumn{2}{c}{1st order differencing}\\
			\cmidrule(lr){2-5}
			& BU Baikal & NAO & BU Baikal & NAO \\
			\midrule
			E-divisive & 1901 1950 & 1924 1960 & 1932 1962 & 1905 1937 \\
			E-divisive+TDA &  1903 1933 & 1903 1933 & 1899 1932 & 1899 1932 \\
			\bottomrule
		\end{tabular}\label{Table 4}
	\end{table}
	
	
	
	\section{Conclusion}\label{Sec:Conclusion}
	
	In this paper we have introduced the emerging concepts of TDA into the CPD framework. Our efforts have been primarily directed at boosting the performance of existing nonparametric CPD algorithms intended to detect changes in distribution, using the analysis of geometry behind the DGP. 
	Under the suggested framework, the CPD detection is performed on a new TDA-derived times series rather than on the original observations.
	The simulation studies carried out using two selected CPD methods - E-Divisive and CvM CPM, have demonstrated the enhanced performance in locating change points when these methods are combined with TDA-based methodology. Notably, such improvements are even more pronounced for multivariate time series whose underlying geometric structure is more complex. Moreover, the new approach shows higher robustness against violations of underlying method assumptions such as independence.
	
	Statistical inference for topological summaries within the TDA framework and asymptotic analysis of Betti numbers in particular, yet remain largely understudied \citep[see the discussion by][references therein]{TDAintro, WassermanTDA, Hiraoka2018}.
    In recent years there have appeared a few papers aiming to investigate asymptotic properties of topological signatures, and among such results are \cite{Owada2018}, \cite{Yogeshwaran2017}, and~\cite{Asym_normality_of_Betti} who focus on limit theorems for Betti numbers for Poisson, binomial and general stationary point processes. To the best of our knowledge, there exist no formal results on the asymptotics of Betti numbers for weakly dependent time series which is the primary focus of this paper. Since theoretical analysis of limiting behavior of Betti numbers constitutes a standalone fundamental mathematical problem falling way beyond the scope of this paper, we leave statistical inference for Betti numbers for weakly dependent time series and associated CPD statistics for future research.
	
	Furthermore, note that due to the invariance property of topological signatures, our current CPD method is insensitive to changes in mean if the variance remains constant. In the future, we plan to investigate TDA approaches allowing to address the detection problem of changes in mean.
	
	The window size $w$ is an important tuning parameter in our new CPD approach. We hypothesize that the window size $w$ is likely to depend on mixing properties of a time series rather than observed sample size $T$. Optimal window size selection can be potentially addressed using subsampling approaches, which constitutes another future research direction.
	
	In addition, we intend to extend the TDA framework to statistical hypothesis testing on change points, namely, to derive TDA-based statistics for change point analysis. Another interesting direction to explore is the impact of dependence, particularly, strong mixing on TDA tools. Finally, we plan to advance the TDA framework to analysis of regime shifts in space-time processes.
	
	\section{Acknowledgement}
	
This work has been partially supported by grants NSF DMS 1925346, NSF DMS 1736368.
and NSF ECCS 1824716. 

\bibliography{references}

\end{document}